\documentclass[10pt,twocolumn,letterpaper]{article}

\usepackage{cvpr}              %

\usepackage[accsupp]{axessibility} %

\usepackage[dvipsnames]{xcolor}

\definecolor{cvprblue}{rgb}{0.21,0.49,0.74}
\usepackage[pagebackref,breaklinks,colorlinks,citecolor=cvprblue]{hyperref}
\usepackage{algorithm}
\usepackage{algorithmic}
\usepackage{amsthm}
\newtheorem{theorem}{Theorem}
\newtheorem{definition}{Definition}

\usepackage{multirow}
\usepackage{booktabs}
\usepackage{amsthm}
\usepackage{amsmath}
\usepackage{bm}
\def\paperID{15354} %
\def\confName{CVPR}
\def\confYear{2024}

\title{Towards Fairness-Aware Adversarial Learning}

\author{Yanghao Zhang\quad Tianle Zhang\quad Ronghui Mu\quad Xiaowei Huang\quad Wenjie Ruan\thanks{Corresponding author}\\
University of Liverpool, UK\\
{\tt\small \{yanghao.zhang, tianle.zhang, ronghui.mu, xiaowei.huang\}@liverpool.ac.uk, w.ruan@trustai.uk}
}

\begin{document}

\maketitle
\begin{abstract}
Although adversarial training (AT) has proven effective in enhancing the model's robustness, the recently revealed issue of fairness in robustness has not been well addressed, i.e. the robust accuracy varies significantly among different categories.
In this paper, instead of uniformly evaluating the model's average class performance, we delve into the issue of robust fairness, by considering the worst-case distribution across various classes.
We propose a novel learning paradigm, named Fairness-Aware Adversarial Learning (FAAL). As a generalization of conventional AT, we re-define the problem of adversarial training as a min-max-max framework, to ensure both robustness and fairness of the trained model.
Specifically, by taking advantage of distributional robust optimization, our method aims to find the worst distribution among different categories, and the solution is guaranteed to obtain the upper bound performance with high probability.
In particular, FAAL can fine-tune an unfair robust model to be fair within only two epochs, without compromising the overall clean and robust accuracies.
Extensive experiments on various image datasets validate the superior performance and efficiency of the proposed FAAL compared to other state-of-the-art methods.
\end{abstract}
    
\section{Introduction}

Deep learning models have undoubtedly achieved remarkable success across various domains, such as computer vision~\cite{russakovsky2015imagenet,zeng2018collaboratively} and natural language processing~\cite{wolf2020transformers}. 
However, they still remain susceptible to deliberate adversarial manipulations of input data~\cite{goodfellow2014explaining,zhang2021fooling,zhang2023generalizing,huang2020survey,huang2023survey,yin2022dimba}. 
Adversarial training techniques~\cite{madry2017towards,wang2022dynamic,jin2022enhancing,jin2023randomized,chen2024nrat} have emerged as a potential solution, aiming to enhance a model's resilience against such vulnerabilities.
These techniques have demonstrated a promising ability to enhance a model's overall robustness, yet the intricate connection between robustness and fairness, as revealed by~\cite{xu2021robust}, demonstrates that the robust accuracy of the models can vary considerably across different categories or classes.
Consider a scenario where an autonomous driving system attains commendable average robust accuracy in recognizing road objects;
despite this success, the system might demonstrate robustness against categories like inanimate objects (\textit{with high accuracy}) while displaying vulnerability to {\em crucial} categories such as ``human" (\textit{with low accuracy}). 
This {\em disparity} or {\em unfair robustness} could potentially {\em endanger} drivers and pedestrians.
Hence, it is vital to ensure consistent, equitable model performance against adversarial attacks by assessing worst-case robustness beyond average levels. 
This provides a more accurate evaluation than the average performance, recognizing the model's limitations while ensuring reliability across diverse categories in real-world applications.

\begin{figure}[!tb]
\centering
\includegraphics[width=0.65\linewidth]{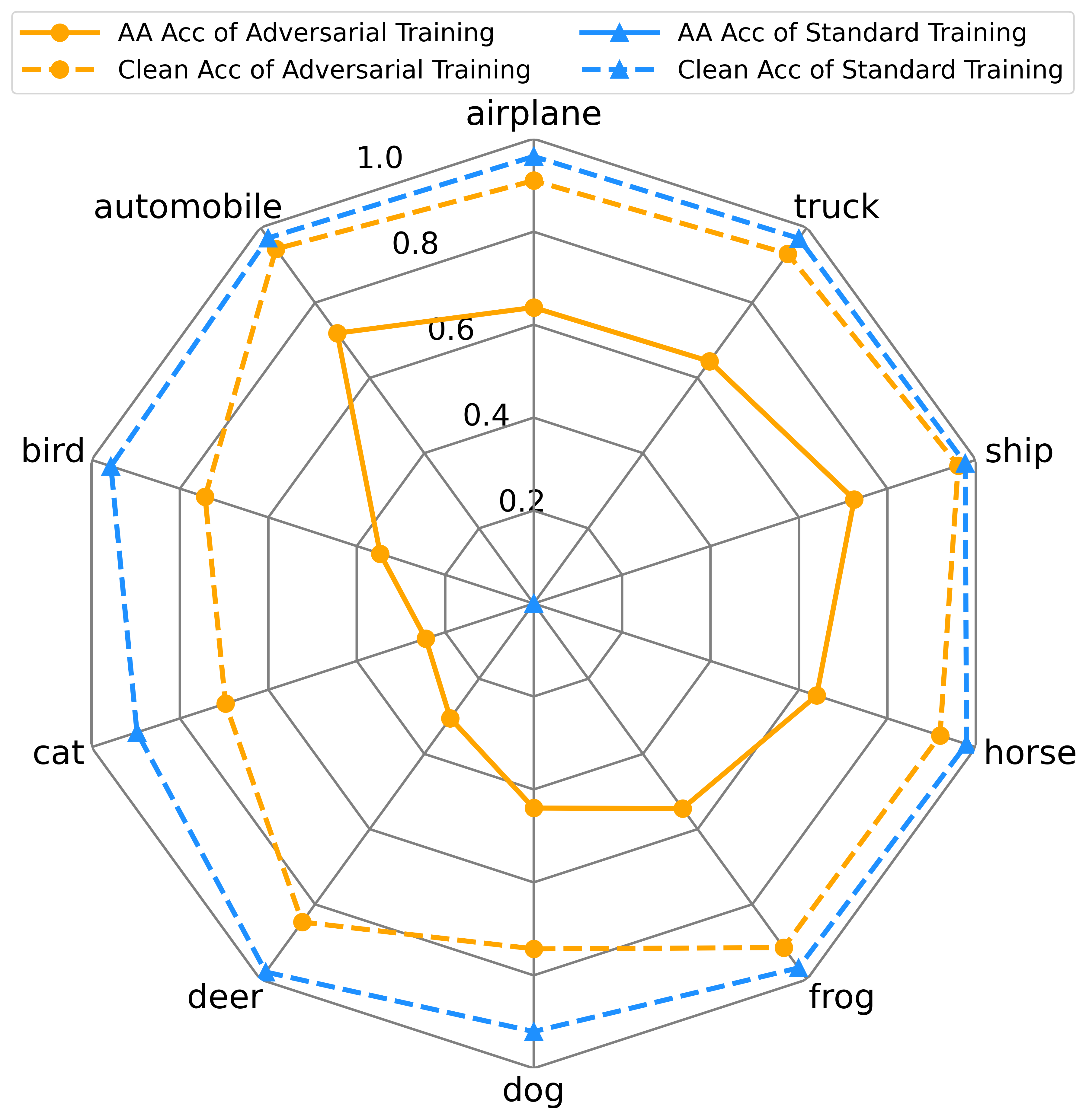}
\caption{Class-wise accuracy of the Wide-ResNet34-10 model on CIFAR-10 dataset, where AA accuracy represents the robust accuracy against AutoAttack.} 
\label{example}
\vspace{-5.5mm}
\end{figure}

\Cref{example} provides an example, displaying the class-wise clean accuracy and robust accuracy against AutoAttack~\cite{croce2020reliable} using the Wide-ResNet34-10~\cite{zagoruyko2016wide} models on CIFAR-10 dataset, where both models are trained via standard training and adversarial training, respectively. 
It comes as no surprise that the model with standard training is vulnerable to adversarial attacks, yet it manages to attain comparable performance across various classes.
In contrast, the adversarially trained model exhibits a noticeable {\em bias}, confidently classifying ``\textit{automobile}" but hesitating with ``\textit{cat}".
That is, the robust accuracy {\em diverges} significantly across classes.
Furthermore, it is noteworthy that even though the class ``\textit{cat}" attains the lowest clean and robust accuracy, the most significant disparity between clean and robust accuracy arises in the case of the class ``\textit{deer}".
This finding highlights an inconsistency between the clean and adversarial performances, where the robust accuracy on different classes illustrates more severe diverges in the model.

This phenomenon is called the ``\textit{robust fairness}" issue, which is first revealed in~\cite{xu2021robust} referring to the gap between average robustness and worst-class robustness.
Recently, some pioneering solutions have been proposed to address the robust fairness issue~\cite{sun2022improving,wei2023cfa,li2023wat}.
These efforts either tackle the robust fairness from the adversarial example generation or adjust the class weights empirically. 
However, essentially, most of these methods can be regarded as instances of \textit{reweighting}, albeit diverse heuristic strategies are adopted.
It is worth noting that such issues are absent in models trained without adversarial training, thus it is obvious that the underlying issue stems from the adversarial perturbations during the adversarial training.
Inspired by but beyond these studies, we are prompted to consider why reweighting strategies are effective in mitigating the robust fairness issue. Intuitively, a model trained adversarially without reweighting fails to achieve high worst-class robust accuracy because it treats all classes equally, yet neglects the fact that the replaced adversarial examples may introduce bias to the final model.
Conversely, reweighting can be perceived as a means of inducing a form of group distribution shift. 
This shift disrupts the uniform optimization of different classes, compelling the model to acquire resistance against these distribution shifts. 
This, in turn, leads to an enhancement in robust fairness.
In line with this, we adopt an alternative approach to address this issue in the adversarial learning paradigm, where the following assumption is made:
\begin{quote}
    {\it The robust fairness issue in the conventional AT is due to the unknown group (class) distribution shift induced by the generated adversarial perturbations, which results in the overfitting problem.}
\end{quote}

As opposed to the heuristic assignment for the reweighting item in the prior works, instead, we are expecting to leverage some optimization techniques for reweighting, such that may bring better results to resolve this overfitting problem.
Hence Distributional Robust Optimization (DRO)~\cite{ben2013robust,duchi2021statistics} naturally emerges as a viable choice.
Rather than assuming a fixed uniform data distribution, DRO acknowledges the inherent distributional uncertainty in real-world data, offering a more resilient and adaptable model structure.
Therefore, this paper delves into the exploration and adaption of DRO, as a sensible solution for the robust fairness challenge.
Specifically, after finding the adversarial example in adversarial training, instead of manually or empirically adjusting the weights for each class, we resort to learning the class-wise distributionally adversarial weights with the pre-defined constraints via DRO.
By learning with these weights, the model will be guided to acquire the capacity to resist unknown group distribution shifts.
The contributions of this paper are summarized as follows:
\begin{itemize}
    \item We investigate the robust fairness issue from the perspective of group/class distributional shift, by taking the recent advances of Distributional Robust Optimization (DRO), which ultimately falls into a reweighting problem. To the best of our knowledge, this work is the first attempt to address the challenge of robust fairness through distributional robust optimization.
    \item We introduce a novel learning paradigm, named FAAL (\underline{\textbf{F}}airness-\underline{\textbf{A}}ware \underline{\textbf{A}}dversarial \underline{\textbf{L}}earning). This innovative approach extends the conventional min-max adversarial training framework into a {\em min-max-max} formulation. The intermediate maximization is dedicated to dealing with the robust fairness issue, by learning with the class-wise distributionally adversarial weights.
    \item Comprehensive experiments are conducted on CIFAR-10 and CIFAR-100 datasets across different models. We empirically validate that the proposed method is able to fine-tune a robust model with intensive bias into a model with {\em both} fairness and robustness within only \textit{two} epochs. %
\end{itemize}

\section{Related Work}

\subsection{Robust Fairness}
It is noted that in traditional machine learning, the definition of \textit{fairness}~\cite{agarwal2018reductions,hashimoto2018fairness} might be different from the \textit{robust fairness} we want to tackle, where the focus of this paper is on mitigating the fairness issue under the scenario against adversarial attacks.

Several works~\cite{xu2021robust,ma2022tradeoff,sun2022improving,wei2023cfa,li2023wat} are explored to alleviate the fairness issue in the robustness.
Xu \etal~\cite{xu2021robust} firstly revealed that the issue of robust fairness occurs in conventional adversarial training, which can introduce severe disparity of accuracy and robustness between different groups of data when boosts the average robustness.
To mitigate this problem, they proposed a Fair-Robust-Learning (FRL) framework, by employing reweight and remargin strategies to finetune the pre-trained model, it is able to reduce the significant boundary error in a certain margin. 
Ma \etal~\cite{ma2022tradeoff} empirically discovered that the trade-off between robustness and robustness fairness exists and AT with a larger perturbation radius will result in a larger variance.
To mitigate the trade-off between robustness and fairness, they add a variance regularization term into the objective function, named FAT, which relieves the trade-off between average robustness and robust fairness.
Sun \etal~\cite{sun2022improving} proposed a method called Balance Adversarial Training (BAT), which adjusts the attack strengths and difficulties of each class to generate samples near the decision boundary for easier and fairer model learning. 
Wei \etal~\cite{wei2023cfa} presented a framework named CFA, which customizes specific training configurations for each class automatically according to which customizes specific training configurations, such that improving the worst-class robustness while maintaining the average performance.
More recently, Li and Liu~\cite{li2023wat} considered the worst-class robust risk, where they proposed a framework named WAT (worst-class adversarial training) and leverage no-regret dynamics to solve this problem.

\subsection{Distributional Robust Optimization}

The origins of DRO are found in the early studies on robust optimization~\cite{ben2009robust,bertsimas2011theory,bennouna2022holistic}, which eventually led to the development of DRO as a tool for handling distributional uncertainties.
The application of DRO to machine learning problems has garnered significant attention like domain generalization~\cite{sagawa2019distributionally}, data distribution shift~\cite{sinha2018certifying,liu2022distributionally}, adversarial robustness~\cite{staib2017distributionally,bui2022unified,bennouna2023certified} and traditional fairness in machine learning~\cite{hashimoto2018fairness,vu2021distributionally,jung2022re,ferry2022improving}.
While the intersection of DRO with robustness and fairness has begun to receive attention, there remain gaps in the literature, particularly in understanding how DRO can address this fairness in an adversarial setting.

\section{Methodology}
\subsection{Preliminaries}

Given the training data drawn from a distribution $P$, when it comes to predicting labels $y \in \mathcal{Y}$ based on input features $x \in \mathcal{X}$,
within a model family denoted as $\Theta$, and utilizing a loss function $ \ell := \Theta \times (\mathcal{X} \times \mathcal{Y}) \rightarrow \mathbb{R}$, 
the conventional training approach for achieving this objective is precisely what's known as empirical risk minimization (ERM):
\begin{equation}
    \min_\theta \mathbb{E}_{(x,y) \sim P} \ \ell(f_\theta(x),y)
\end{equation}
In traditional adversarial training, the focus is on identifying the worst-case perturbation for each input. This is formulated as a min-max problem, as defined in~\cite{madry2017towards}. Mathematically, it can be expressed as follows:
\begin{equation}
\label{eq_at}
    \min_\theta \mathbb{E}_{(x,y) \sim P} \max_{\delta \in B_\epsilon} \ \ell(f_\theta(x+\delta),y)
\end{equation}
When accounting for class fairness, specifically the performance across different classes, \cref{eq_at} can be rewritten as:
\begin{equation}
\label{eq_at_fair}
    \min_\theta \frac{1}{C}\sum_{c=1}^C   \mathbb{E}_{(x,y) \sim P_c} \max_{\delta \in B_\epsilon} \ \ell(f_\theta(x+\delta),y)
\end{equation}
It is noted that when a batch contains an equal number of data points for each class, \cref{eq_at_fair} is technically identical to \cref{eq_at}.
According to the definition of distributional robust optimization~\cite{ben2013robust,duchi2021statistics}, we now consider minimizing the expected loss in the worst-case scenario over a set of uncertain distributions. This can be mathematically expressed as:
\begin{equation}
    \min_\theta \sup_{Q \in \mathcal{Q}} \mathbb{E}_{(x,y) \sim Q} \max_{\delta \in B_\epsilon} \ \ell(f_\theta(x+\delta),y)
\end{equation}
where $\mathcal{Q}$ represents the uncertainty set, encompassing the range of potential test distributions for which we seek the model to exhibit commendable performance aligned with the data distribution $P$.

To establish the connection between robust fairness and DRO, we can naturally delineate the classes as distinct groups within the training data. Subsequently, the uncertainty set $\mathcal{Q}$ can be defined with respect to these groups.
Specifically, the setting of group DRO are borrowed~\cite{hu2018does,oren2019distributionally,sagawa2019distributionally}, where the training distribution $P$ is assumed to be a mixture of $C$ groups (classes) $P_C$ indexed by $c = \{1, 2, . . . , C\}$.
Thus the uncertainty set $\mathcal{Q}$ is defined as any mixture of these classes, \ie $\mathcal{Q}:=\{\sum_{c=1}^C q_cP_c: \bm{q} \in \Delta_C\}$, where $\Delta_C$ is the probability simplex.
Hence, the worst-case risk can be reformulated as the most detrimental  combination across different groups, taking into account the expected loss for each class:
\begin{equation}
\label{groupdro}
    \min_\theta \sup_{\bm{q} \in \Delta_C} \sum_{c=1}^C q_c \cdot \mathbb{E}_{(x,y) \sim P_c} \max_{\delta \in B_\epsilon} \ \ell(f_\theta(x+\delta),y)
\end{equation}
\\
However, as proven in~\cite{hu2018does}, applying DRO directly to robust learning training is overly pessimistic, which often yields results that do not surpass those achieved by a classifier adversarially trained using ERM.
This outcome can be attributed to the specific classification loss function and the distributions that DRO seeks to encompass for the purpose of robustness are notably extensive.
A similar pattern of failure is also encountered in the context of group DRO~\cite{sagawa2019distributionally}, and they advocate that sufficient regularization is required for over-parameterized neural networks to enhance worst-group generalization.

\subsection{Problem Definition}
By disentangling the \cref{groupdro}, it becomes evident that it can be interpreted as a \textit{reweighting} objective of the ERM framework within the context of adversarial settings, incorporating with the weighted factor $\bm{q}$.
Empirical evidence has demonstrated the efficacy of several reweighting methods~\cite{xu2021robust,benz2021robustness,wei2023cfa} in improving robust fairness and all of them fall into the same paradigm of \cref{groupdro}.
This aligns with the assumption stated in the introduction section, while
we seek to leverage some optimization techniques for promoting fairness directly, rather than a heuristic assignment.

As previously mentioned, the extensive range of distributions encompassed by the uncertainty set $\mathcal{Q}$ could present difficulties for DRO in sustaining its robustness. Nevertheless, the pivotal factor in addressing the group DRO lies in configuring the uncertainty set.
To tackle this issue, we advocate an alternative solution: instead of relying solely on substantial regularization~\cite{sagawa2019distributionally}, we propose to use a straightforward yet effective ambiguity set with extra constraint. 
This is achieved by defining the ambiguity set as $\mathcal{Q}':=\{\sum_{c=1}^C q_cP_c: d(\mathcal{U}, \bm{q})\leq \tau, \bm{q}\in \Delta_C\}$, where $d(\cdot,\cdot)$ represents some distance metrics measuring the difference between two distributions, $\tau$ is the constraint parameter and $\mathcal{U}$ is the uniform distribution.
This choice of $\mathcal{Q}'$ shrinks the width of $\mathcal{Q}$ and allows us to learn models that are robust to some group shifts, rather than identically uniform distribution among different classes.
Hence, our final objective in addressing the challenge posed by potential group distribution shifts within an adversarial setting can be denoted as:
\vspace{-3.5mm}
\begin{equation}
\label{faal}
    \min_\theta \max_{d(\mathcal{U}, \bm{q} ) \leq \tau, \bm{q}\in \Delta_C} \sum_{c=1}^C q_c \cdot \mathbb{E}_{(x,y) \sim P_c} \max_{\delta \in B_\epsilon} \ \ell(f_\theta(x+\delta),y)
\end{equation}
Since the robust fairness issue occurs in general adversarial training, we also do not know the real class distribution shift may occur at test time, especially under adversarial training.
The uncertainty set $\mathcal{Q}'$ encodes the possible test distributions that we want our model to perform well on.  
Therefore, a suitable divergence ball around the class distribution confers robustness to a set of distributional shifts. 
In our settings, we use KL divergence as $d$ in our following experiments, as we will see the KL-DRO~\cite{bennouna2023certified} has its unique property which provides the optimal solution for handling the fairness issue. 
In other words, such an overall objective will optimize the worse distribution of the neighborhood around the uniform distribution for different classes by learning those adversarial examples.

\begin{algorithm}[!h]
\caption{Fairness-Aware Adversarial Learning}
\label{faal_algorithm}
\textbf{Input}: Training set $\{X,Y\}$, total epochs $T$, adversarial radius $\epsilon$, step size $\alpha$, the number of adversarial iteration $K$, model $f$  parameterized by $\theta$, the number of mini-batches~$M$, batch size $B$, distribution shift constraint $\tau$\\
\textbf{Output}: A robust and fair model
\begin{algorithmic}[1] %
\FOR{$t = 1 \ldots T$}
\FOR{ $i = 1 \ldots M$}
\STATE  \# {\it Phase 1: Inner maximization}\\$\delta=0$
\FOR{ $j = 1 \ldots K$} 
\STATE $\delta = \delta + \alpha \cdot {\rm sign}(\nabla_\delta \ell_{\rm CE}(f_\theta(x_i+\delta),y_i))$
\STATE $\delta = {\rm max}({\rm min}(\delta, \epsilon),-\epsilon)$ 
\ENDFOR
\STATE $x_{i}^{adv} = {\rm clip}(x_{i}+\delta, 0, 1)$ 
\\ 
\STATE \# {\it Phase 2: Intermediate maximization}\\ $\ell_i = \ell_{\rm CE}(f_\theta(x_{i}^{adv}),y_i,{\rm reduction=}$`${\rm none}$'$) $ \\ 
\# Calculate the cross-entropy loss for each instance\\
\FOR{ $c = 1 \ldots C$}
\STATE $\ell'_c = \ell_{\rm CW}(f_\theta(x_{i}^{adv},y_i)[y_i=c] $ 
\\ \# Calculate the average margin for each class $c$
\ENDFOR
\STATE $\bm{w^{cda}_*} = {\tt solve\_kl\_dro}(\bm{\ell'},\tau)$
\\ \# Solve the optimal class-wise weights for the current batch under the worst distribution via DRO
\STATE $\mathcal{L}_{\rm FAAL} = \frac{1}{B}\sum_{i=1}^B \bm{w^{cda}_*}[y] \cdot \ell_i \cdot C $
\STATE \# {\it Phase 3: Outer Minimization} \\$\theta=\theta-\nabla_{\theta}  \mathcal{L}_{\rm FAAL}$  
\ENDFOR
\ENDFOR
\STATE \textbf{return} Robust model $f_\theta$ with high fairness
\end{algorithmic}
\end{algorithm}

\subsection{Fairness-Aware Adversarial Learning }
Based on the above objective, we propose a novel adversarial learning paradigm, named \underline{\textbf{F}}airness-\underline{\textbf{A}}ware \underline{\textbf{A}}dversarial \underline{\textbf{L}}earning (FAAL)\footnote{Our code is available at https://github.com/TrustAI/FAAL}, to improve the robust fairness via distributional robust optimization.
Specifically, within the intermediary stage of the conventional adversarial training, \ie between inner maximization and outer minimization, we introduce a class-wise distributionally adversarial weight for orientating the learning directions among different categories, which can be optimally solved by leveraging distributional robust optimization.
By incorporating this weight into the outer minimization process to update the model's parameters, the class (group) distribution shift can be protected to alleviate the robust fairness issue.

To provide a clearer illustration of the whole learning problem, we break down it into three distinct stages:
\begin{itemize}
    \item \textit{Phase 1}: Inner maximization for finding adversarial examples;
    \item \textit{Phase 2}: Intermediate maximization for finding the distributionally adversarial weight (worst-case distribution around the uniform distribution);
    \item \textit{Phase 3}: Outer minimization for updating model's parameters.
\end{itemize} 
\textit{Phases} \textit{1} and \textit{3} are the classic processes of conventional adversarial training, and \textit{Phase 2} is the core element of our proposed learning paradigm, as we assume that tackling the unknown class distributional shift can contribute to enhancing robust fairness.
The whole procedure is summarized in \cref{faal_algorithm}.
In the following content, we replace the notion of $\bm{q}$ in \cref{faal} with $\bm{w^{cda}}$ for convenience and define it as Class-wise Distributionally Adversarial Weight.
\begin{definition}
[CDAW: Class-wise Distributionally Adversarial Weight]
Given a class-wise objective loss $\ell'_c \in \mathbb{R} $ on the adversarial examples, for all classes $c \in C$, the optimal Class-wise Distributionally Adversarial Weight vector $\bm{w^{cda}_*}$ aims to maximize the overall loss:
\vspace{-5mm}
\begin{equation}
\begin{split}
\mathcal{L}_{\rm FAAL}:= & \max \sum_{c=1}^{C} w^{cda}_c\ell'_c \\
 \text{s.t.} \quad & d(\mathcal{U},\bm{w^{cda}} ) \leq \tau, \bm{w^{cda}} \in \Delta_C
\end{split}
\label{faal_eq1}
\end{equation}

\begin{equation}
\bm{w^{cda}_*} := \arg\max \mathcal{L}_{\rm FAAL}
\label{cdaw1}
\end{equation}
\end{definition}

In the case of a reweighting strategy employed to address robust fairness, it dictates the learning trajectory for each individual class. 
By optimizing the model using this optimized weight, the model is exposed to learning from the worst-case distribution under the pre-defined constraint $\tau$, such that fairness among different classes will be encouraged.
When $\bm{w^{cda}}$ is identical to $\mathcal{U}$, \ie $\tau=0$, it reduces the regular mean calculation for the overall loss, making the entire learning paradigm regress to conventional adversarial training that contains \textit{Phases 1} and \textit{3} only.
We use KL divergence as $d$ in the intermediate maximization, such that it can be solved via the \textit{conic convex optimization}, and another elegant property of it on the generalization can be obtained, as demonstrated in \cref{theorem} below.

\begin{theorem}
\label{theorem}
Given the loss $\mathcal{L}_{\rm FAAL}$ in \cref{faal_eq1} on the observed distribution, and suppose the regular loss $\mathcal{L}=:\frac{1}{C} \sum_{c=1}^{C}\ell'_c$ on the test distribution with unknown group distribution shift,
then the following holds for all $\bm{w^{cda}} \in \Delta_C$:
\begin{equation}
    {\rm Pr}(\mathcal{L}_{\rm FAAL}\textgreater\mathcal{L}) \geq 1-e^{-\tau n+O(n)}
\end{equation}
Where ${\rm KL}(\mathcal{U},\bm{w^{cda}})\leq \tau$, $\mathcal{U}$ is the uniform distribution. 
\end{theorem}
\Cref{theorem} tells that the $\mathcal{L}_{\rm FAAL}$ is guaranteed to be the upper bound of $\mathcal{L}$ with high probability given the large number of observed sample $n$.
In line with this, it enjoys a strong generalization where the performance on the test distribution with some unknown group distribution shift is at least as good as the estimated performance with high probability.
So the solution of the class-wise distributionally adversarial weight solving by the convex optimization is optimal and will provide protection on the unknown class distribution shift.
As cross-entropy loss cannot well-represents how good the discrepancy between classes~\cite{cao2019learning}, we instead use the CW margin loss~\cite{carlini2017towards} as $\mathcal{L}$ for calculating the class-wise distributionally adversarial weight:
\begin{equation}
    \ell'_c := \mathbb{E}_{(x,y) \sim P_c}(\max_{j!=y}{z_j} - z_y)
\end{equation}
where $z_j$ is the probability of the class $j$, \ie the softmax output of the network.
It is noted that the objective functions in \textit{Phases 1-3} of our learning paradigm are not necessarily consistent, so the proposed learning mechanism is flexible and can be combined with any min-max adversarial approaches.
In the following experiments, we will mainly solve the distributional robust optimization on the bounded margin loss among classes, which provides better performance than using cross-entropy loss in the intermediate maximization. 
More details can be found in the Appendix.

\begin{table*}[t]
\centering
\caption{Evaluation of different fine-tuning methods on CIFAR-10 dataset using WRN-34-10 model. The best result is highlighted in \textbf{Bold}.}
\label{results_wrn}
\resizebox{0.75\linewidth}{!}{
\renewcommand\arraystretch{1.3}
\begin{tabular}{lccccc}
\toprule[1.5pt]
\multirow{2}{*}{Adversarially Trained WRN-34-10 Model}                  & \multirow{2}*{\shortstack{Fine-Tuning\\Epochs}}& \multicolumn{4}{c}{Average Accuracy (Worst-class Accuracy) (\%)}     \\ \cline{3-6}           
&& Clean &  PGD-20 & CW-20 & AutoAttack\\
\midrule[1pt]
PGD-AT    &-&   86.07 (69.70)  &  55.90 (29.90)  & 54.29 (28.30) &    52.46 (24.40) \\
\quad +  Fine-tune with FRL-RWRM$_{ 0.05}$&80 & 83.25 (\textbf{74.80})  & 50.37 (38.10)  & 49.77 (36.60) & 46.97 (33.10)    \\ 
\quad +   Fine-tune with FRL-RWRM$_{0.07}$&80 & 85.12 (71.60)& 52.56 (37.10) & 51.92 (35.50) & 49.60 (31.70)   \\ 
\quad + Fine-tune with ${\rm FAAL_{AT}}$ &\textbf{2}& \textbf{86.23} (69.70)& 54.00 (37.60) &53.11 (36.90) & 50.81 (35.70)\\
\quad + Fine-tune with ${\rm FAAL_{AT-AWP}}$ &\textbf{2}& 85.47 (69.40)& \textbf{56.46} (\textbf{39.20}) &\textbf{54.50} (\textbf{38.10}) & \textbf{52.47} (\textbf{36.90})
\\

\midrule

TRADES &- & 84.92 (67.00)&55.32 (27.10)  &53.92 (24.80)  &\textbf{52.51} (23.20) \\
\quad + Fine-tune with FRL-RWRM$_{ 0.05}$ &80 & 82.90 (72.70) & 53.16 (40.60)  & 51.39 (36.30) &49.97 (\textbf{35.40}) \\
\quad + Fine-tune with FRL-RWRM$_{0.07}$ &80 & 85.19 (70.90)  & 53.76 (39.20)  & 52.92 (36.80) &51.30 (34.60)\\

\quad + Fine-tune with ${\rm FAAL_{AT}}$&\textbf{2} & \textbf{85.96} (\textbf{75.00})& 53.46 (39.80) &52.72 (38.20) & 50.91 (35.30)\\
\quad + Fine-tune with ${\rm FAAL_{AT-AWP}}$ &\textbf{2}& 85.39 (72.90)& \textbf{56.07} (\textbf{43.30}) &\textbf{54.16} (\textbf{38.60}) & 52.45 (\textbf{35.40})\\

\midrule
MART &-& 83.62 (67.90) & \textbf{56.22} (32.50)  & \textbf{52.79} (25.70) & \textbf{50.95} (22.00) \\ 
\quad +  Fine-tune with FRL-RWRM$_{ 0.05}$ &80& \textbf{83.72 }(\textbf{71.80}) & 52.16 (37.50) & 50.73 (35.00) & 49.19 (31.70)
\\
\quad + Fine-tune with FRL-RWRM$_{0.07}$ &80& 82.09 (\textbf{71.80}) & 50.86 (36.00) & 49.78 (33.00) &47.78 (30.30)
\\

\quad + Fine-tune with ${\rm FAAL_{AT}}$ &\textbf{2}& 83.49 (68.00)& 51.65 (37.80) &50.36 (37.10) & 48.63 (34.00)\\

\quad + Fine-tune with ${\rm FAAL_{AT-AWP}}$&\textbf{2} & 82.17 (64.00)& 54.31 (\textbf{39.50}) &51.72 (\textbf{37.70}) &50.31 (\textbf{36.40}) 

\\

\midrule[1pt]
TRADES-AWP&- &       85.35 (67.90)    &     59.20 (28.80)  &             \textbf{57.14} (26.50)       &  \textbf{ 56.18} (25.80)    \\
\quad + Fine-tune with FRL-RWRM$_{ 0.05}$&80 & 82.31 (65.90)  & 49.90 (31.70) & 49.68 (34.00) & 46.50 (27.70)
\\
\quad + Fine-tune with FRL-RWRM$_{0.07}$&80 & 84.24 (65.70) & 48.63 (30.90) &  49.77 (31.50) & 46.53 (28.60)
\\

\quad + Fine-tune with ${\rm FAAL_{AT}}$&\textbf{2} & 87.02 (\textbf{76.30})& 52.54 (35.00)& 51.70 (34.40) & 49.87 (30.60)\\

\quad + Fine-tune with ${\rm FAAL_{AT-AWP}}$ &\textbf{2}& \textbf{86.75} (74.80)& \textbf{57.14}(\textbf{43.40})& 55.34 (\textbf{40.10}) &  53.93 (\textbf{37.00})

\\

\bottomrule[1.5pt]
\end{tabular}
}
\end{table*}

\section{Experimental Results}

Given our method's focus on the robust fairness challenge, it is reasonable to assume that the model already possesses a certain degree of average robustness. 
Otherwise, considering the robust fairness issue might not yield meaningful results.
Hence, the question arises: \textit{Is it imperative to initiate the training of a model from the beginning for achieving fairness with a certain robustness level?}
In the next section, we first test our approach through adversarial fine-tuning, and then explore if training from scratch with our method can offer additional benefits.

\subsection{Fine-tuning for Enhancing Robust Fairness}
\textbf{Baselines and experiment settings}:
We first conducted experiments on CIFAR-10 dataset~\cite{krizhevsky2009learning}, which is popularly used for adversarial training evaluation.
We use the average \& worst-class accuracy under different adversarial attacks (Clean / PGD~\cite{madry2017towards} / CW~\cite{carlini2017towards} / AutoAttack~\cite{croce2020reliable}) as the evaluation metrics. 
The perturbation budget is set to $\epsilon=8/255$ on CIFAR-10 dataset.
FRL~\cite{xu2021robust} is the only existing state-of-the-art technique from the recent literature which performs fine-tuning to a pre-trained model for improving robust fairness.
FRL proposed two strategies based on TRADES~\cite{zhang2019theoretically} for enhancing robust fairness, including reweight (RW) and remargin (RM).
Hence we apply the best versions of FRL from their paper: FRL-RWRM with ${\tau_1=\tau_2=0.05}$ and FRL-RWRM with ${\tau_1=\tau_2=0.07}$, where $\tau_1$ and $\tau_2$ are the fairness constraint parameters for reweight and remargin of FRL, we name them FRL-RWRM$_{0.05}$ and FRL-RWRM$_{0.07}$ for short.
The results of FRL are reproduced using their public code, where the target models are fine-tuned for 80 epochs and the best results are presented.

In terms of the proposed method, although we utilize the PGD-AT adversary method by default (named FAAL$_{\rm AT}$ for short), it is completely compatible with other AT approaches like TRADES or MART. To achieve this, one just needs to keep the original implementation for both inner maximization and outer minimization unchanged and add the intermediate maximization independently.
We found that 2 epochs of fine-tuning are enough to improve the robust fairness greatly, without sacrificing too much average clean/robust accuracy.
We set the value of $\tau$ in our method as $0.5$, and the learning rate is configured from $0.01$ in the first epoch and drops to $0.001$ in the second epoch. 

\begin{figure*}[!h]
\centering
\includegraphics[width=0.76\textwidth]{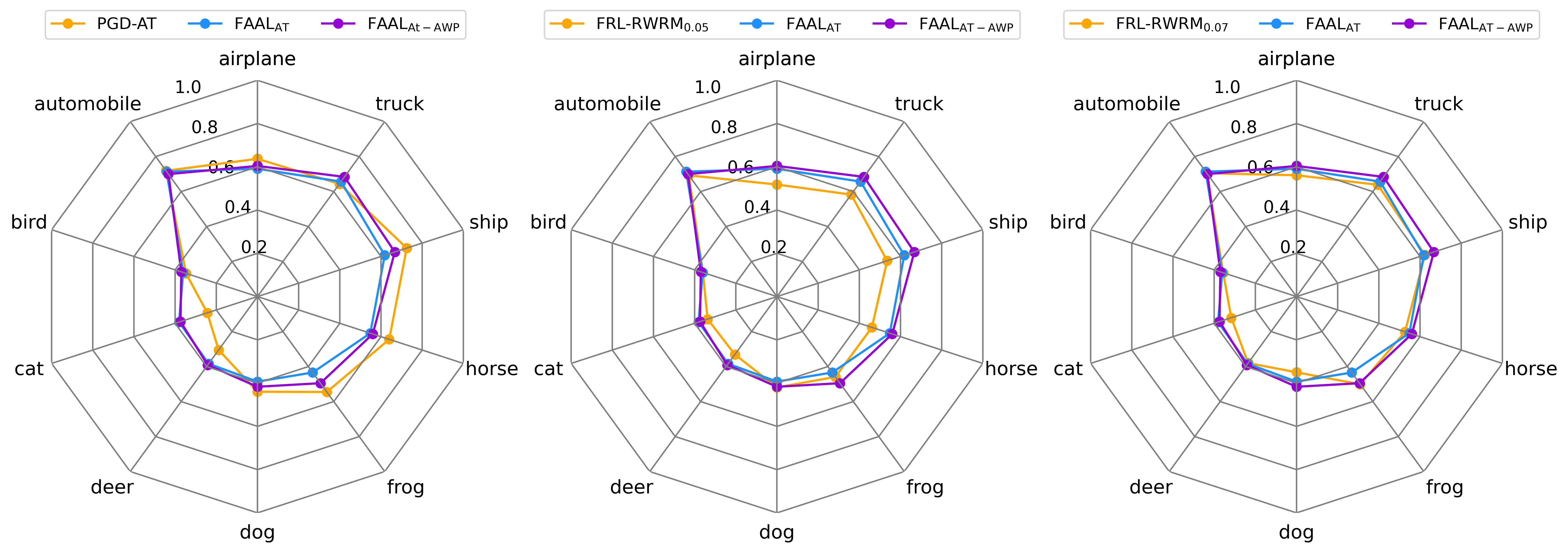}
\caption{Class-wise robust accuracy against AutoAttack after fine-tuning the PGD adversarially trained WRN model} 
\label{pre_ft}
\end{figure*}
 
\Cref{results_wrn} demonstrates our main results of finetuning Wide-Resnet34-10 (WRN-34-10) models~\cite{zagoruyko2016wide} on CIFAR-10 dataset, where different state-of-the-art adversarial defensed methods are adopted,
including PGD-AT~\cite{madry2017towards}, TRADES~\cite{zhang2019theoretically}, MART~\cite{wang2019improving} and AWP~\cite{wu2020adversarial}.
We can see that FAAL$_{\rm AT}$ outperforms the two FRL methods with respect to both average and worst-class robustness. 
Notably, in the majority of cases, FAAL$_{\rm AT}$ achieves this without significant compromises in clean accuracy, unlike FRL methods which tend to trade off the clean accuracy for improving robustness. 
FAAL$_{\rm AT}$ promotes the worst-class AutoAttack (AA) accuracy by approximately $2.6\%$ than FRL for fine-tuning PGD-AT, MART, and AWP.
Except for fine-tuning TRADES, both methods yield comparable performance, this is partially due to that FRL is a TRADES-based method and it takes advantage of knowing the source method.
Most importantly, FRL requires many epochs (80 epochs) to obtain the best results, while our method, is able to achieve better results within \textbf{only 2 epochs}.
As adversarially training a large model with high robustness is already time-consuming, to circumvent the need for retraining the model from the beginning, FAAL offers a solution for saving time and computational resources. 
It demonstrates the capability to quickly fine-tune a robust model that initially lacks fairness, resulting in a model that is both robust and fair. 
Due to the space limit, similar improvements on the Preact-Resnet model can be found in the Appendix. 

\noindent \textbf{Strong adversarial attacks can help?}
The remargin of FRL~\cite{xu2021robust} claims that increasing the perturbation margin can help for obtaining better robust fairness, while this may hurt the average clean accuracy/robustness, as indicated in \cref{results_wrn}.
Certainly, there is a \textit{trade-off} existing between the average robustness and worst-class robustness, but is it necessary to increase the perturbation margin $\epsilon$ for improving the class-wise robustness? 
We question whether this is a mandatory requirement for improvements, and we assume the benefits come from the stronger strength of adversarial perturbation.
Hence, instead of enlarging the perturbation margin, we capitalize on the flexibility of our learning framework and
integrate our method with AWP~\cite{wu2020adversarial}, a well-regarded model weight perturbation technique, to strengthen the attacks.
As shown in \cref{results_wrn}, when combining with AWP, FAAL$_{\rm AT-AWP}$ further enhances the worst-class robust accuracy on WRN34-10 models, 
especially for the original one adversarially trained with AWP.
FAAL$_{\rm AT-AWP}$ is almost unharmed on the improvement to the original unfair models most of the time. 
Therefore, it is not compulsory to enlarge the perturbation margin to gain better results, where applying a stronger adversary indeed benefits robust fairness without enlarging the perturbation margin.
\Cref{pre_ft} visualizes the results of class-wise AA accuracy for the comparison of the proposed method FAAL$_{\rm AT}$ and FAAL$_{\rm AT-AWP}$, and two FRL baselines, respectively.
It can be seen that FAAL boost the worst-class robust accuracy, presenting outstanding capacity to improve robust fairness with high effectiveness and efficiency, respectively, where it outperforms FRL not only for the average/worst-class robustness but also for the very rare fine-tuning steps.

\begin{figure*}[!h]
\centering
\includegraphics[width=0.76\textwidth]{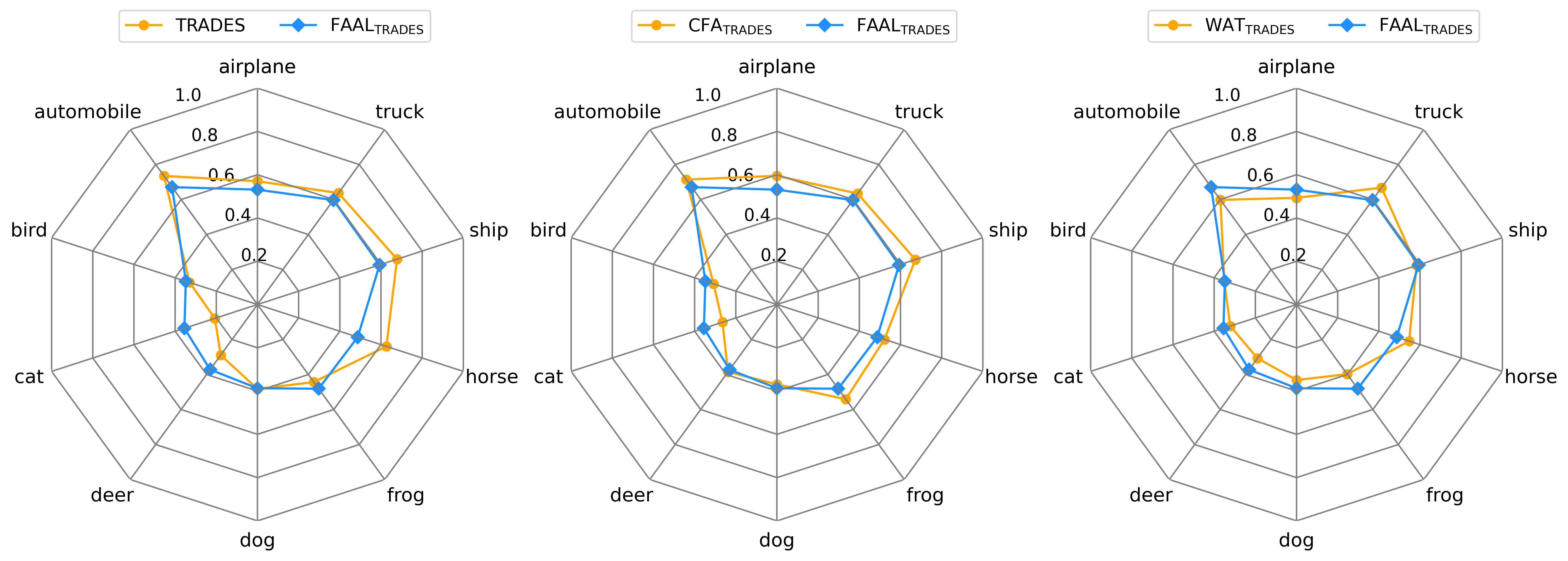}
\caption{Class-wise robust accuracy against AutoAttack after adversarially trained PRN-18 model from scratch} 
\label{train200_plot}
\end{figure*}

\begin{table}[!h]
\centering
\caption{Training from scratch with different methods on CIFAR-10 dataset using Preact-ResNet18 model. }
\label{train200_prn}
\resizebox{\linewidth}{!}{
\renewcommand\arraystretch{1.5}
\begin{tabular}{lcc}
\toprule[1.5pt]
\multirow{2}{*}{Adversarially Trained PRN-18 Model}           & \multicolumn{2}{c}{Average Acc (Worst-class Acc) (\%)}     \\ \cline{2-3}         &Clean  &  AutoAttack
\\
\hline
PGD-AT   &  \textbf{82.72} (55.80)& 47.38 (12.90)   \\ 
TRADES &  82.54 (66.10) & 49.05 (20.70)    \\ 
CFA$_{\rm AT}$ & 80.82 (64.60) & \textbf{50.10} (24.40)
\\
CFA$_{\rm TRADES}$ & 80.36 (66.20)  & \textbf{50.10} (26.50)
\\
WAT$_{\rm TRADES}$ & 80.37 (66.00) & 46.16 (30.70)
\\
FAAL$_{\rm AT}$ &82.20 (62.90)  & 49.10 (\textbf{33.70})\\

FAAL$_{\rm TRADES}$ & 81.62 (\textbf{68.90})   &
48.48 (33.60)
\\

\bottomrule[1.5pt]
\end{tabular}
}
\end{table}

\subsection{Training from Scratch for Enhancing Fairness}

Previous sections demonstrate the effectiveness and efficiency of the proposed approaches. 
Here we also investigate the advancements by training the model from the ground up using our method.
We compare our methods with two common adversarial training methods (PGD-AT~\cite{madry2017towards} and TRADES~\cite{zhang2019theoretically}) and two recent state-of-the-art techniques: CFA~\cite{wei2023cfa} and WAT~\cite{li2023wat}, which have been proposed to mitigate the robust fairness issues recently.
We adversarially trained Preact-ResNet-18 models~\cite{he2016deep} for 200 epochs with a learning rate of 0.1, which will be decayed by a factor of 0.1 at 100 and 150 epochs, successively.
We start to facilitate the proposed intermediate maximization (see \cref{faal_algorithm} lines 9-14) after the 100-th epoch with the only hyper-parameter $\tau$ from 0.25 and enlarge it to 0.5 after the 150-th epoch.
In addition, similar to CFA using weight average, we also applied EMA~\cite{izmailov2018averaging,wang2022self}, to gain a more stable performance, however, we only applied it after the 100-th epoch, where we start to apply the intermediate maximization.
We report the best results under AutoAttack on the average accuracy and worst-class accuracy in \cref{train200_prn}.
Besides, \cref{train200_plot} visualizes the results of different training approaches including the proposed FAAL$_{\rm TRADES}$
with other 3 TRADES-based models: TRADES, CFA$_{\rm TRADES}$ and WAT$_{\rm TRADES}$ respectively.
We can observe that FAAL outperforms other approaches on the worst-class clean/robust accuracy, with less sacrifice on the average robustness.

\begin{table}[!h]
\centering
\caption{Result comparison of different methods on CIFAR-100 dataset using ResNet18 model.}
\label{train100_cifar100}
\resizebox{\linewidth}{!}{
\renewcommand\arraystretch{1.5}
\begin{tabular}{lcc}
\toprule[1.5pt]
\multirow{2}{*}{Adversarially Trained RN-18 Model}           & \multicolumn{2}{c}{Average Acc (Worst-class Acc) (\%)}     \\ \cline{2-3}         &Clean  &  AutoAttack
\\
\hline
TRADES &  54.57 (19.00) & 23.57 (1.00)    \\ 
\quad +  Fine-tune with FRL-RWRM$_{ 0.05}$& 52.55 (22.00) &21.11 (2.00)
\\
\quad +  Fine-tune with FAAL$_{\rm AT}$ &\textbf{58.50} (21.00)  & 21.91 (2.00)
\\
\quad +  Fine-tune with FAAL$_{\rm AT-AWP}$ &58.41 (19.00)  & 23.44 (2.00)
\\
\quad +  Fine-tune with FAAL$_{\rm TRADES}$ & 54.96 (18.00) & 22.71 (2.00)
\\
\quad +  Fine-tune with FAAL$_{\rm TRADES-AWP}$ & 54.90 (18.00) & 23.25 (2.00)
\\
\midrule
CFA$_{\rm TRADES}$ & 55.57 (\textbf{23.00})  & \textbf{24.56} (2.00)
\\
WAT$_{\rm TRADES}$ & 53.99 (19.00) & 22.89 (\textbf{3.00})
\\
FAAL$_{\rm AT}$  & 56.84 (16.00)& 21.85 (\textbf{3.00})
\\

FAAL$_{\rm TRADES}$ &55.87 (21.00) &23.57 (\textbf{3.00})
\\
\bottomrule[1.5pt]
\end{tabular}
}
\end{table}
\subsection{Additional Experiments on CIFAR-100 dataset}
The experiments above mainly focused on CIFAR-10 dataset, which only has 10 classes in the dataset.
In this section, we explore the proposed FAAL into a more challenging dataset, \ie CIFAR-100 with 100 categories.
Similarly, we reported the results of the average/worst clean accuracy and AutoAttack accuracy.
The value of $\tau$ in our method is set to 0.05.
For fine-tuning, we compare our proposed method FAAL with FRL-RWRM$_{0.05}$, it can be seen that FAAL is able to achieve comparable to FRL-RWRM while reducing the amount of learning epoch up to 40 times (2 epochs \vs 80 epochs).
For full adversarial training, following the experimental settings in WAT~\cite{li2023wat}, we train the ResNet-18 models for 100 epochs via different adversarial training approaches, where the learning rate is decayed from 0.1 to 0.01 and 0.001 at the 75-th epoch and the 90-th epoch, respectively.
We compare the results of FAAL compared to three baselines, \ie TRADES, CFA and WAT.
It can be seen in \cref{train100_cifar100} that FAAT$_{\rm TRADES}$ achieves the highest worst-class robust accuracy (same as WAT), but it remains comparable results on the average robustness without sacrificing the average/worst-class clean accuracy. 
More details of the training settings can be found in the Appendix.

\begin{table}[t]
\centering
\caption{Comparison among different SOTA methods, all models are trained with the same number of samples under a single NVIDIA 3090Ti GPU in the same conda environment.}
\label{compution_time}
\resizebox{\linewidth}{!}{
\renewcommand\arraystretch{1.5}
\begin{tabular}{lccccc}
\toprule[1.5pt]
\multirow{3}{*}{Methods}            &\multicolumn{2}{c}{Training time per epoch (\textit{min})}  & \multirow{3}{*}{\shortstack{Reweighting\\ level}} &  \multirow{3}{*}{\shortstack{Adversary \\free}}  &  \multirow{3}{*}{\shortstack{Validation \\set}}  \\ \cline{2-3}          &\multirow{2}{*}{\shortstack{CIFAR-10 \\(PRN-18)}}  &  \multirow{2}{*}{\shortstack{CIFAR-100 \\ (RN-18) }}
\\
\\
\hline
TRADES &  2.63 & 2.68& fixed  & $\times$ & $\times$
\\
FRL-RWRM& 2.73 & 2.80& epoch & $\times$ & $\checkmark$\\
WAT & 2.88& 3.00 & epoch & $\times$& $\checkmark$
\\
CFA & 2.75 & 2.78 & epoch & $\checkmark$ & $\checkmark$
\\
FAAL  & 2.69 & 2.73& batch & $\checkmark$ & $\times$
\\
\bottomrule[1.5pt]
\end{tabular}
}
\end{table}

\section{Essential Differences to SOTAs}

In this section, we highlight the essential differences of FAAL with existing state-of-the-art works, including FRL~\cite{xu2021robust}, WAT~\cite{li2023wat} and CFA~\cite{wei2023cfa}. 
Both FRL and WAT are {\em TRADES-based} approaches, which require a {\em separate} validation set for performing the reweight strategies. 
For example, FRL updates the lagrangian multiplier according to the performance of the validation set to meet the fairness constraints, so it requires many epochs for fine-tuning since it needs to search the whole space to achieve the optimal equilibrium without fairness constraint violation.
Also, the remargin strategy in FRL essentially sacrifices some average clean accuracy. We argue that it is not necessary to enlarge the margin for improvement, which can be achieved by combining stronger perturbations instead.
As another TRADES-based variant, WAT leverages no-regret dynamics and also relies on the validation set to tune the class weights for the current epoch training, 
Similarly, CFA proposed to apply the weight averaging only if the performance on the extra validation set meets a certain threshold, and relies on empirically adjusting the class margins and class regularization based on the performance of the previous epoch.

Different form those methods that rely on historical performance or an extra validation set for manual or heuristic weight adjustment per class in each epoch, our method {\em bypasses} these requirements.
FAAL introduces an additional conic convex optimization problem after the adversarial example generation, based solely on the current batch's objective loss, the bringing solving cost is negligible. 
The comparison of training computation time and other key properties is illustrated in \cref{compution_time}.
As model training can be unpredictable due to random mini-batch sampling, causing quick shifts in class distribution and bias that may differ from previous epochs or validations.
More importantly, FAAL can generalize to {\em any} adversarial training methods, as our intermediate maximization is a completely independent component plugged into the popular min-max framework, so it is not limited to any adversaries, unlike some methods FRL and WAT that are restricted to TRADES variants.
Our data-driven component enhances flexibility in managing the balance between average robustness and robust fairness during adversarial training, and demonstrates its potential in handling various distribution shifts for the current batch.

\section{Conclusion}

In conclusion, we establish a connection between robust fairness and potential overfitting issues caused by the unknown group distribution shift, and present a new fairness-aware adversarial learning paradigm to address robust fairness via distributional robust optimization. 
Compared to state-of-the-art methods, extensive experiments on CIFAR-10 and CIFAR-100 datasets demonstrate the effectiveness and superior efficiency of the proposed approach. 
Notably, by just two epochs of fine-tuning, our training strategy can transform a biased robust model into one with high fairness with little cost on average accuracy.
We believe our research provides a meaningful contribution to the discourse on robustness and fairness in machine learning, deepening our insight into the model's behaviors under adversarial settings.

\section*{Acknowledgement}
The research is supported by the UK EPSRC under project EnnCORE [EP/T026995/1] and University of Liverpool.

\clearpage
{
    \small
    \bibliographystyle{ieeenat_fullname}
    \bibliography{main}
}

\end{document}